\UseRawInputEncoding
\documentclass[conference]{IEEEtran}
\IEEEoverridecommandlockouts
\usepackage{cite}
\usepackage[numbers,sort&compress]{natbib}
\usepackage{hyperref}
\usepackage{amsmath,amssymb,amsfonts}
\usepackage{algorithmic}
\usepackage{graphicx}
\usepackage{textcomp}
\usepackage{xcolor}
\usepackage{multirow}
\usepackage{array}
\usepackage{listings}
\def\BibTeX{{\rm B\kern-.05em{\sc i\kern-.025em b}\kern-.08em
    T\kern-.1667em\lower.7ex\hbox{E}\kern-.125emX}}
\begin{document}

\title{OTTER: Open-Tagging via Text-Image Representation for Multi-modal Understanding}

\author{
\begin{minipage}{0.9\textwidth} % 控制宽度，0.9\textwidth 可调
\centering
Jieer Ouyang\textsuperscript{1}, Xiaoneng Xiang\textsuperscript{1}, Zheng Wang\textsuperscript{1}, Yangkai Ding\textsuperscript{1}\\
\textsuperscript{1}Huawei Singapore Research Center, Singapore \\
\texttt{ouyang.jieer@huawei.com, xiangxiaoneng@u.nus.edu, wangzheng155@huawei.com, dingyangkai@huawei.com}
\end{minipage}
}

\maketitle

\begin{abstract}
We introduce OTTER, a unified open-set multi-label tagging framework that harmonizes the stability of a curated, predefined category set with the adaptability of user-driven open tags. OTTER is built upon a large-scale, hierarchically organized multi-modal dataset, collected from diverse online repositories and annotated through a hybrid pipeline combining automated vision-language labeling with human refinement. By leveraging a multi-head attention architecture, OTTER jointly aligns visual and textual representations with both fixed and open-set label embeddings, enabling dynamic and semantically consistent tagging. OTTER consistently outperforms competitive baselines on two benchmark datasets: it achieves an overall F1 score of 0.81 on Otter and 0.75 on Favorite, surpassing the next-best results by margins of 0.10 and 0.02, respectively. OTTER attains near-perfect performance on open-set labels, with F1 of 0.99 on Otter and 0.97 on Favorite, while maintaining competitive accuracy on predefined labels. These results demonstrate OTTER's effectiveness in bridging closed-set consistency with open-vocabulary flexibility for multi-modal tagging applications.
\end{abstract}


\section{Introduction}
In today's information dense digital systems, individuals and organizations continuously generate, store, and interact with heterogeneous media ranging from images to documents across a variety of personal and collaborative platforms. The ability to retrieve and organize such content efficiently is no longer a convenience but a fundamental requirement for productivity, knowledge management, and decision-making. Central to this capability is the presence of a coherent, semantically meaningful tagging and labeling system that can bridge the gap between raw data and human-interpretable concepts. Recent advances have demonstrated that well-structured annotation schemes can dramatically improve search precision~\cite{zhang2022research,jeya2022content,mcauley2012image}. In particular, open-ended tagging architectures, capable of assigning descriptive, context-aware labels beyond a fixed vocabulary, offer the flexibility to capture nuanced semantics in dynamic environments. By enabling accurate and adaptive annotation across modalities, such systems empower users to transform unstructured media into accessible, reusable knowledge assets, thereby enhancing both personal information management and large-scale content curation.

\begin{figure}[h]
    \centering
    \includegraphics[width=\columnwidth]{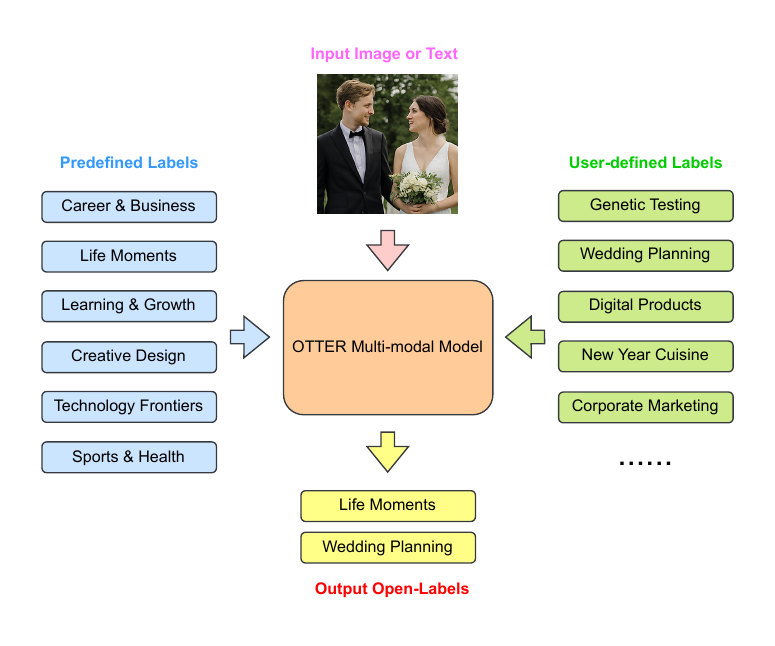}
    \caption{Illustration of the OTTER framework. An input image, a wedding photo, is tagged with both a fixed category (Life Moments) and a personalized open label (Wedding Planning). OTTER enables accurate assignment within a predefined category space while flexibly accommodating user-specific custom tags.}
    \label{fig:otter-demo}
\end{figure}

Multi-label tagging, which assigns multiple relevant categories to a single image or text instance, has been extensively studied but remains challenging due to label co-occurrence, semantic overlap, and the need to capture both fine-grained and abstract concepts~\cite{yang2016exploit,wang2017multi,gao2021learning,you2020cross,chen2019learning,ye2020attention}. Open-label multi-label tagging extends coverage to unseen classes via pretrained visual or textual encoder with strong generalization capability, yet these typically assume a certain constraint label space, limiting adaptability to emerging categories~\cite{ridnik2023ml,timmermann2025lm,zhang2022tip}. Some recent multi-modal large language model approaches~\cite{abdelhamed2024you,han2024onellm,xu2022open,ali2023clip} have improved flexibility and coverage, but still face persistent challenges: ensuring semantic precision for fine-grained categories and effectively grounding textual labels in visual evidence.

\begin{table*}[]
\small
\setlength{\tabcolsep}{8pt}
\renewcommand{\arraystretch}{1.3}
\caption{Predefined Text-Image Tags with Examples}
\label{tab:predefined-tags}
\begin{tabular}{
>{\centering\arraybackslash}m{0.15\textwidth}|
>{\centering\arraybackslash}m{0.55\textwidth}|
>{\centering\arraybackslash}m{0.2\textwidth}
}
\hline
\textbf{Predefined Tags} & \textbf{Tag Definition} & \textbf{Example Scenarios} \\
\hline\hline
\begin{minipage}[t]{\linewidth}\raggedright Career \& Business\end{minipage} &
\begin{minipage}[t]{\linewidth}\raggedright Encompasses market analysis, corporate strategy, financial management, human resources, product development, marketing strategies, supply chain management, customer relations, entrepreneurship, and international business.\vspace{3pt}\end{minipage} &
\begin{minipage}[t]{\linewidth}\raggedright Annual reports, financial statements,industry trends.\end{minipage} \\
\hline
\begin{minipage}[t]{\linewidth}\raggedright Life Moments\end{minipage} &
\begin{minipage}[t]{\linewidth}\raggedright Captures moments from personal life, including family activities, travel experiences, culinary exploration, holiday celebrations, and personal achievements—reflecting the richness and diversity of daily living.\vspace{3pt}\end{minipage} &
\begin{minipage}[t]{\linewidth}\raggedright Food and entertainment, holidays, important schedules, computer games.\end{minipage} \\
\hline
\begin{minipage}[t]{\linewidth}\raggedright Creative Design\end{minipage} &
\begin{minipage}[t]{\linewidth}\raggedright Focuses on creative thinking, design theory, art appreciation, case studies in design, and analysis of popular trends—aimed at sparking creativity and providing design inspiration.\vspace{3pt}\end{minipage} &
\begin{minipage}[t]{\linewidth}\raggedright Outfit design, fashion design.\end{minipage} \\
\hline
\begin{minipage}[t]{\linewidth}\raggedright Learning \& Growth\end{minipage} &
\begin{minipage}[t]{\linewidth}\raggedright Includes educational news, learning resources, skill training, personal development strategies, career planning, and self-improvement—designed to support lifelong learning and personal growth.\vspace{3pt}\end{minipage} &
\begin{minipage}[t]{\linewidth}\raggedright Career advancement, error notebooks.\end{minipage} \\
\hline
\begin{minipage}[t]{\linewidth}\raggedright Sports \& Health\end{minipage} &
\begin{minipage}[t]{\linewidth}\raggedright Covers healthy eating, fitness, mental health, disease prevention, wellness habits, and medical information—aimed at promoting a healthy lifestyle and improving quality of life.\vspace{3pt}\end{minipage} &
\begin{minipage}[t]{\linewidth}\raggedright Health check reports, fitness activities, medical treatments.\end{minipage} \\
\hline
\begin{minipage}[t]{\linewidth}\raggedright Tech Frontiers\end{minipage} &
\begin{minipage}[t]{\linewidth}\raggedright Content related to the latest technological inventions, innovations, scientific research progress, future trend forecasts, and practical applications—especially breakthrough technologies that may have a significant impact on industries or society.\vspace{3pt}\end{minipage} &
\begin{minipage}[t]{\linewidth}\raggedright Electrical vehicles, artificial intelligence, digital products.\end{minipage} \\
\hline
\end{tabular}
\end{table*}

To address these limitations, we propose a unified open-set multi-label tagging framework, OTTER, that harmonizes the stability of a curated, predefined category set with the adaptability of user-driven open tags. Our approach is grounded in the construction of a large-scale, hierarchically organized multi-modal dataset, sourced from diverse online repositories and annotated through a hybrid pipeline that combines automated vision-language labeling with human refinement. This dataset serves as the foundation for training an open-tagging model that fuses visual and textual representations, and enables dynamic alignment between input content and both fixed and open label embeddings. The training regime is designed to mirror real-world tagging conditions: fixed labels are interleaved with true open-set labels and sampled negative open-set candidates. Based on our design, as shown in Fig.~\ref{fig:otter-demo}, when presented with a user-supplied image or text snippet, such as a wedding photograph accompanied by a brief description, the system can accurately assign both a stable category (Life Moments) and a user-specific open tag (Wedding Planning), thereby supporting consistent organization while preserving the expressive granularity needed for personalized retrieval.

Our contribution can be summarized as follows:
\begin{itemize}
\item We design a dual-layer tagging paradigm that integrates a stable taxonomy of predefined categories with an extensible open-tag layer, enabling simultaneous consistency in annotation and adaptability to evolving user needs.
\item We create a large-scale, hierarchically structured multi-modal dataset, curated from heterogeneous online sources and annotated through a synergistic combination of automated vision-language models and expert human validation, ensuring both breadth of coverage and semantic reliability.
\item We build up an efficient tagging architecture that leverages multi-head attention to align and process multi-modal inputs with both fixed and open-set label embeddings, achieving high accuracy in open-tagging multi-label classification.
\end{itemize}

\section{Related Work}
\subsection{Multi-Label Tagging}
Multi-label image or text tagging, which assigns multiple relevant categories to a single instance, remains a challenging topic due to label co-occurrence and semantic overlap. Some early research apply region-based methods to localize discriminative areas before classification, either by leveraging bounding box annotations~\cite{yang2016exploit} or by recurrently discovering attentional regions~\cite{wang2017multi,gao2021learning}. More recent designs integrate cross-modality attention with semantic graph embeddings~\cite{you2020cross} to capture fine-grained cues. While these approaches can improve per-label accuracy, they often suffer from coarse or redundant region proposals, high computational overhead, and difficulty in localizing abstract or scene-level concepts. Some other methods model inter-label dependencies to improve prediction consistency, such as semantic-specific graph representations~\cite{chen2019learning} and attention-driven dynamic graph convolutional networks~\cite{ye2020attention}. However, these correlation-based models can be sensitive to dataset bias, overfitting to frequent co-occurrences and propagating spurious relations.

\begin{table*}[]
\small
\setlength{\tabcolsep}{8pt}
\renewcommand{\arraystretch}{1.3}
\caption{Predefined Text-Image Tags and Sub-Tags}
\label{tab:predefined-subtags}
\begin{tabular}{
>{\centering\arraybackslash}m{0.15\textwidth}|
>{\centering\arraybackslash}m{0.78\textwidth}
}
\hline
\textbf{Predefined Tags} & \textbf{Sub-tags} \\
\hline\hline
\begin{minipage}[t]{\linewidth}\raggedright Career \& Business\end{minipage} &
\begin{minipage}[t]{\linewidth}\raggedright Market Analysis, Corporate Strategy, Financial Management, Human Resources, Product Development, Marketing Strategies, Supply Chain Management, Customer Relations, Entrepreneurship, International Business, Annual Reports, Financial Statements, Industry Trends, Data Analysis\vspace{3pt}\end{minipage} \\
\hline
\begin{minipage}[t]{\linewidth}\raggedright Life Moments\end{minipage} &
\begin{minipage}[t]{\linewidth}\raggedright Everyday Life, Family Life, Travel Experiences, Culinary Exploration, Festival Celebrations, Daily Life, Food and Entertainment, Travel Guides, Game Screenshots, WeChat Chat Screenshots, Cards and Certificates\vspace{3pt}\end{minipage} \\
\hline
\begin{minipage}[t]{\linewidth}\raggedright Creative Design\end{minipage} &
\begin{minipage}[t]{\linewidth}\raggedright Inspirational Design, Creative Thinking, Artworks, Artwork Appreciation, Design Case Studies, Fashion Trends, Design Inspiration, Outfit Design, Fashion Design, Illustration Styles, 3D Art\vspace{3pt}\end{minipage} \\
\hline
\begin{minipage}[t]{\linewidth}\raggedright Learning \& Growth\end{minipage} &
\begin{minipage}[t]{\linewidth}\raggedright Learning and Growth, Educational Information, Learning Resources, Skills Training Posters, Career Planning, Self-Improvement, Workplace Advancement, Error Books, Exam Papers, Homework Questions, Attending Classes, Work Skills\vspace{3pt}\end{minipage} \\
\hline
\begin{minipage}[t]{\linewidth}\raggedright Sports \& Health\end{minipage} &
\begin{minipage}[t]{\linewidth}\raggedright Sports and Healthcare, Healthy Diet Therapy, Sports and Fitness, Mental Health, Disease Prevention, Healthy Habits, Medical Information, Healthy Lifestyles, Improving Quality of Life, Medical Examination Reports, Disease Treatment\vspace{3pt}\end{minipage} \\
\hline
\begin{minipage}[t]{\linewidth}\raggedright Tech Frontiers\end{minipage} &
\begin{minipage}[t]{\linewidth}\raggedright Frontiers of Technology, Technological Inventions, Technological Innovations, Scientific Research Progress, Technology Forecasts, Technology Application Cases, Breakthrough Technologies, New Energy Vehicles, Artificial Intelligence, Digital Products, New Scientific Discoveries, Internet and Cybersecurity, Aerospace and Advanced Manufacturing, Life Sciences and Medical Technology, Clean Energy and Sustainable Development, Military Technology\vspace{3pt}\end{minipage} \\
\hline
\end{tabular}
\end{table*}

\subsection{Open-Label Tagging}
Open-label multi-label tagging extends the task to unseen categories by leveraging auxiliary semantic information. Early zero-shot methods adapted closed-set models~\cite{yang2016exploit,ridnik2023ml} or learned cross-modal compatibility functions between images and label embeddings, as in attribute-based label embedding~\cite{akata2015label} or principal-direction ranking in word-vector space~\cite{zhang2016fast}. Bilinear attention networks~\cite{kim2018bilinear} and convex combinations of semantic embeddings~\cite{norouzi2013zero} further enriched image-text interactions. Many of these approaches rely heavily on textual embeddings without sufficient visual grounding, which can hinder generalization to unseen categories. Recent work has shifted toward multi-modal large language models and unified vision-language frameworks.~\cite{abdelhamed2024you} employs MLLMs to generate rich textual descriptions from images, fusing them with visual features for zero-shot classification, reducing the need for dataset-specific prompt engineering.~\cite{zhang2022tip} adapt large pretrained encoders to training-free adapters for few-shot scenarios.~\cite{han2024onellm} aligns multiple modalities to language within a single framework, enabling cross-modal reasoning beyond vision-text pairs. Open-vocabulary multi-label classification with dual-modal decoders~\cite{xu2022open} aligns visual and textual features for better generalization, while~\cite{ali2023clip} integrates multi-modal CLIP embeddings with attention-based heads for zero-shot multi-label prediction.~\cite{timmermann2025lm} adapts positive asymmetric loss to mitigate long-tail bias in open-vocabulary settings. These methods significantly improve flexibility and coverage, they still face challenges in ensuring semantic precision for fine-grained categories, and maintaining efficiency when the candidate label set is extremely large. 

\section{Methodology}
We introduce a two-tier tagging framework that combines a stable set of predefined categories with flexible open tags, enabling both consistent annotation and user-driven personalization. We construct a large, diverse, and hierarchically organized multi-modal dataset annotated through a blend of automated vision-language labeling and human refinement to train a model that fuses visual and textual features via multi-head attention for open-set multi-label classification. The training strategy mirrors real-world conditions by mixing fixed labels with sampled negative open-set labels and probabilistically including true open-set labels.

\subsection{Task Formulation}\label{AA}
In contemporary digital environments, individuals routinely store and interact with diverse media such as images, documents, and articles across personal devices including smartphones, tablets, and laptops. Efficient retrieval and organization of such content critically depend on the availability of a coherent and semantically meaningful tagging system. While ad-hoc or unstructured tagging can offer short-term convenience, it often leads to inconsistencies and reduced retrieval performance over time. Therefore, before assigning tags to user content, it is essential to design a principled and extensible tag taxonomy that balances coverage, usability, and personalization.

To address this need, we propose a two-tier tag taxonomy comprising predefined tags and open tags. predefined tags are predefined based on categories that reflect the most common types of content users encounter in their daily digital interactions. These categories, detailed in Table~\ref{tab:predefined-tags}, are designed to cover the majority of images and textual materials stored on personal devices. By providing a stable and standardized set of tags, we aim to ensure interoperability, reduce cognitive load, and facilitate consistent annotation practices across different users.

\begin{figure*}[t]
  \centering
  \includegraphics[width=\textwidth]{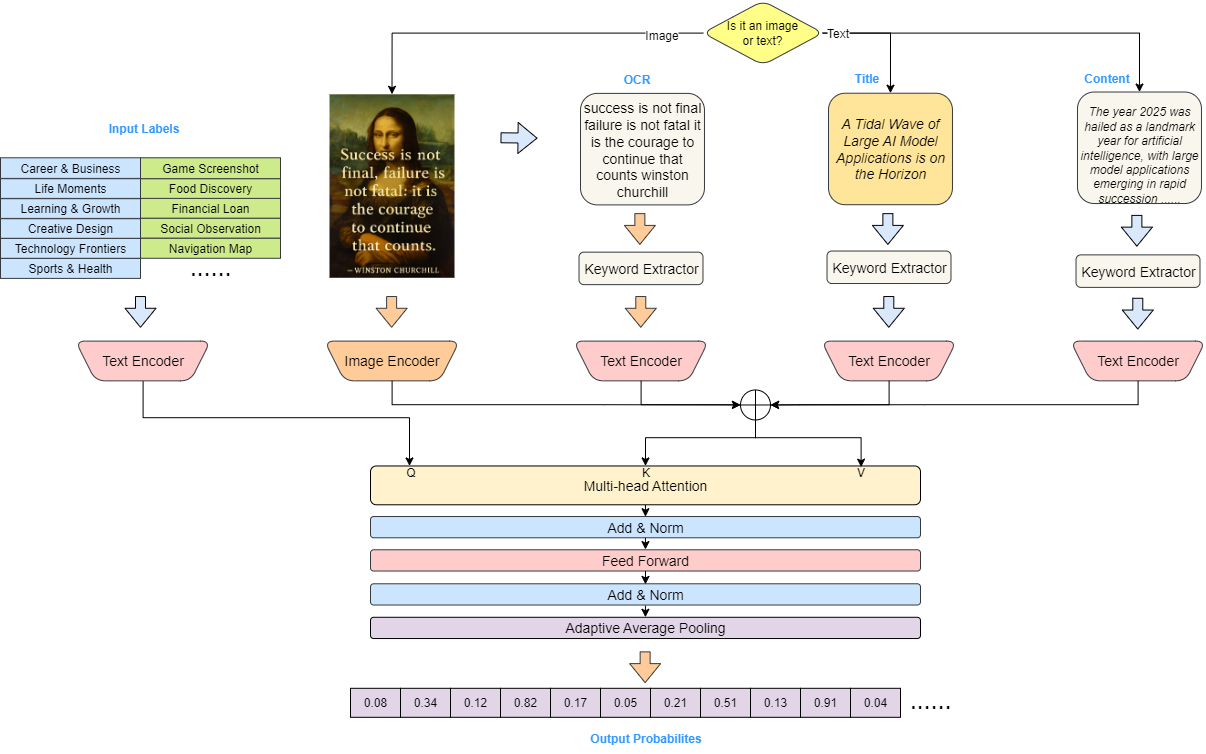}
  \caption{Illustration of OTTER model architecture. The model employs a multi-head attention mechanism in which fixed and open-set label embeddings, encoded by a shared text encoder, serve as queries attending over fused visual and textual features derived from images or text inputs. Visual features from a vision backbone and textual features from OCR-based keyword extraction or direct text processing are aligned in a shared embedding space, summed to form keys and values, and processed through attention, adaptive average pooling, and a sigmoid layer to yield independent label probabilities.}
  \label{fig:otter-model}
\end{figure*}

However, recognizing that user interests and preferences are inherently diverse and dynamic, our taxonomy also incorporates open tags. These allow users to define custom labels that reflect their unique contexts, projects, or thematic needs. For example, a user preparing for a wedding might create a dedicated “Wedding Planning” tag, enabling them to consolidate all related materials—such as venue images, guest lists, and design inspirations—under a single, easily retrievable category. This hybrid approach aligns with recent advances in user-centered and adaptive information system design, which emphasize personalization and flexibility \cite{fleury2024multi}.

By combining the stability of preset tags with the flexibility of open tags, our system seeks to optimize both retrieval efficiency and personal relevance. This formulation not only addresses the challenges of large-scale personal content management but also lays the groundwork for adaptive tagging strategies that can evolve alongside user behavior and emerging content types.

\subsection{Dataset Construction}
To ensure that our model can accurately recognize tags representing users' daily life activities, it is essential to prepare a training dataset with broad coverage across diverse scenarios. Prior research has emphasized that the representativeness and diversity of training data are critical for robust multi-modal recognition systems~\cite{guo2024multimodal,wang2020makes,li2024data}. Our dataset construction process follows a two-stage pipeline: (1) acquiring a sufficiently large and diverse corpus of candidate data, and (2) annotating this data with both predefined and open-ended tags.

In the first stage, we decomposed each high-level tag category into a set of fine-grained sub-tags, as outlined in Table~\ref{tab:predefined-subtags}. This hierarchical decomposition strategy has been shown to improve label granularity and downstream classification performance in multi-modal tagging tasks. For each sub-tag, we collected a substantial number of relevant images and associated textual descriptions from online sources, ensuring coverage across different visual styles, contexts, and linguistic expressions. This approach aligns with recent dataset construction practices that leverage web-scale retrieval to maximize domain coverage while maintaining semantic specificity.

In the second stage, we employed a large language-vision model to assign both fixed predefined tags and dynamically generated open tags to each image-text pair (see Appendix~\ref{appendix} for the annotation commands). Automated annotation with large models has been increasingly adopted in multi-modal dataset pipelines to accelerate labeling and capture nuanced semantic attributes~\cite{xie2024medtrinity}. To further enhance label quality, we conducted manual verification, cleaning, and balancing of the dataset. This post-processing step ensured that the distribution of samples across tags was as uniform as possible, mitigating class imbalance issues that can adversely affect model generalization.

By combining hierarchical tag decomposition, large-scale web data acquisition, automated multi-modal annotation, and human-in-the-loop refinement, our dataset construction methodology provides a scalable and high-quality foundation for training models capable of accurately recognizing a wide spectrum of user daily life tags.

\subsection{Model Architecture}
Our framework adopts a multi-head attention mechanism that explicitly computes the attention between open-set label embeddings and the image or text embedding extracted from the input material, as decoupling label queries from image or text features allows for flexible and efficient multi-label classification across diverse label spaces~\cite{ridnik2023ml}. This design enables the model to flexibly attend to relevant semantic cues in the input, regardless of whether the label space is predefined or dynamically extended. As illustrated in Fig.~\ref{fig:otter-model}, both fixed and open-set label descriptions are first transformed into dense representations via a shared text encoder~\cite{liu2019roberta}, which serves as the query input to the attention module. For image inputs, a vision backbone~\cite{dosovitskiya2010animageis,guo2024multimodal,tschannen2025siglip} is employed to obtain high-level visual features. In parallel, any embedded text within the image is extracted using an OCR pipeline, followed by keyword extraction based on statistical or graph-based ranking algorithms~\cite{ramos2003using,zhang2024graph}. The resulting keywords are then encoded using the same text encoder applied to labels. For purely textual inputs, keywords of both the document title and body are extracted, and the resulting content is encoded into the shared embedding space based on the same text encoder. The encoded visual and textual features are summed element-wise to form the key and value inputs to the multi-head attention layer~\cite{ali2023clip}. This additive fusion ensures that both modalities contribute equally to the attention computation, allowing the model to capture complementary cues. The attended feature maps are subsequently aggregated via adaptive average pooling, and a sigmoid activation is applied to produce independent probability estimates for each candidate label.

\subsection{Training Strategy}
\begin{figure}[h]
    \centering
    \includegraphics[width=\columnwidth]{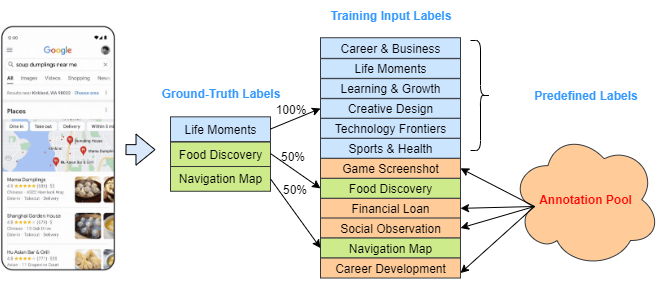}
    \caption{Illustration of OTTER training strategy. Training data are constructed by combining six fixed predefined labels with a set of open-set labels drawn from a global label pool, where sampled labels differ from the ground truth and serve as negative examples alongside the true labels in the input.}
    \label{fig:otter-data}
\end{figure}

To ensure that the trained model can effectively address challenges encountered in real-world deployments, the training procedure is designed to closely mimic the conditions of the target application domain. As illustrated in Fig.~\ref{fig:otter-data}, all open-set labels present in the training corpus are first aggregated into a comprehensive open-set label pool. Each training instance—whether an image or a text sample—typically has two to three annotated ground-truth labels. For each training step, the input label set is composed of six fixed, predefined labels and a variable number of open-set labels. The predefined labels remain constant across the entire training process, while the open-set labels are sampled from the label pool. These sampled labels acting as negative samples, together with the ground-truth open-set labels acting as positive samples, form the training input labels. To further enhance training diversity and robustness, the ground-truth open-set labels are not always included in the input, but injected with a certain probability. During optimization, the parameters of the image encoder and text encoder are frozen while the rest are tuned.

\section{Experiments}
\subsection{Experimental Setup}
\subsubsection{Dataset and Ground Truth} We conduct experiments on two proprietary multi-modal datasets, Otter and Favorite, designed to evaluate both predefined and open-set label recognition. The Otter dataset is constructed in a manner analogous to our training corpus: a collection of image-text and label pairs was crawled from the web according to a predefined set of sub-labels, ensuring coverage across diverse semantic categories. To obtain open annotations, we employed the LLM assisted labeling pipeline, in which open labels were generated automatically. The Favorite dataset comprises image-text and label pairs collected from multiple real users in diverse daily-life scenarios, thereby reflecting authentic and heterogeneous content distributions. Unlike Otter, the Favorite dataset was annotated entirely through manual labeling by trained annotators, following a standardized protocol to ensure inter-annotator agreement. Together, these datasets provide complementary evaluation settings: Otter offers large-scale, LLM-assisted annotations aligned with predefined label taxonomies, while Favorite captures real-world variability and open-set conditions through human-verified labels. This dual-dataset design enables a comprehensive assessment of model performance across both controlled and unconstrained multi-modal recognition scenarios.

\subsubsection{Baselines} We benchmark OTTER against a diverse set of state-of-the-art multimodal baselines. The selected baselines include CLIP~\cite{guo2024multimodal}, SigLIP2~\cite{tschannen2025siglip}, Qwen2VL~\cite{wang2024qwen2}, and WhatDoYouSee~\cite{abdelhamed2024you}. This selection covers a representative spectrum of architectural paradigms, from dual-encoder contrastive models to instruction-tuned MLLMs, enabling a comprehensive comparative analysis.

\begin{table*}[t]
\caption{Performance Comparison on Otter and Favorite Datasets with Predefined Tag and Open-Tag Settings}
\centering
\vspace{-3mm}
\label{tab:otter_favorite_tags}
\setlength{\tabcolsep}{3pt}
\begin{tabular}{
    l|
    >{\centering\arraybackslash}p{0.9cm}
    >{\centering\arraybackslash}p{0.9cm}
    >{\centering\arraybackslash}p{0.9cm}|
    >{\centering\arraybackslash}p{0.9cm}
    >{\centering\arraybackslash}p{0.9cm}
    >{\centering\arraybackslash}p{0.9cm}|
    >{\centering\arraybackslash}p{0.9cm}
    >{\centering\arraybackslash}p{0.9cm}
    >{\centering\arraybackslash}p{0.9cm}|
    >{\centering\arraybackslash}p{0.9cm}
    >{\centering\arraybackslash}p{0.9cm}
    >{\centering\arraybackslash}p{0.9cm}
}
\hline
\multirow{3}{*}{Method} 
& \multicolumn{6}{c|}{Otter} & \multicolumn{6}{c}{Favorite} \\ \cline{2-13}
& \multicolumn{3}{c|}{Predifined Tag} & \multicolumn{3}{c|}{Open Tag} & \multicolumn{3}{c|}{Predifined Tag} & \multicolumn{3}{c}{Open Tag} \\ \cline{2-13}
& Prec. & Rec. & F1 & Prec. & Rec. & F1 & Prec. & Rec. & F1 & Prec. & Rec. & F1 \\ \hline
CLIP         & 0.58 & \textcolor{red}{0.77} & \textcolor{blue}{0.66} & 0.78 & 0.83 & 0.80 & 0.57 & 0.71 & 0.63 & 0.79 & 0.81 & \textcolor{blue}{0.80} \\
SIGLIP       & 0.40 & 0.65 & 0.50 & 0.59 & 0.78 & 0.67 & 0.53 & 0.61 & 0.57 & 0.61 & 0.68 & 0.64 \\
Qwen2VL      & \textcolor{blue}{0.68} & 0.49 & 0.57 & \textcolor{blue}{0.89} & \textcolor{blue}{0.91} & \textcolor{blue}{0.90} & \textcolor{blue}{0.67} & \textcolor{red}{0.73} & \textcolor{red}{0.70} & 0.76 & \textcolor{blue}{0.83} & 0.79 \\
WhatDoYouSee & 0.43 & 0.65 & 0.52 & 0.73 & 0.82 & 0.77 & 0.68 & 0.49 & 0.57 & \textcolor{blue}{0.82} & 0.68 & 0.74 \\
OTTER        & \textcolor{red}{0.70} & \textcolor{blue}{0.65} & \textcolor{red}{0.67} & \textcolor{red}{0.98} & \textcolor{red}{1.00} & \textcolor{red}{0.99} & \textcolor{red}{0.77} & \textcolor{blue}{0.70} & \textcolor{blue}{0.69} & \textcolor{red}{0.94} & \textcolor{red}{1.00} & \textcolor{red}{0.97} \\ \hline
\end{tabular}
\end{table*}

\begin{table}[htbp]
\caption{Performance Comparison on Otter and Favorite datasets}
\centering
\vspace{-3mm}
\label{tab:otter_favorite}
\setlength{\tabcolsep}{3pt} % 缩小列间距
\begin{tabular}{l|
    >{\centering\arraybackslash}p{0.9cm}
    >{\centering\arraybackslash}p{0.9cm}
    >{\centering\arraybackslash}p{0.9cm}|
    >{\centering\arraybackslash}p{0.9cm}
    >{\centering\arraybackslash}p{0.9cm}
    >{\centering\arraybackslash}p{0.9cm}}
\hline
\multirow{2}{*}{Method} & \multicolumn{3}{c|}{Otter} & \multicolumn{3}{c}{Favorite} \\ \cline{2-7} 
 & Prec. & Rec. & F1 & Prec. & Rec. & F1 \\ \hline
CLIP           & 0.62 & \textcolor{red}{0.82} & \textcolor{blue}{0.71} & 0.61 & \textcolor{blue}{0.75} & 0.67 \\
SIGLIP         & 0.46 & 0.76 & 0.57 & 0.58 & 0.66 & 0.62 \\
Qwen2VL        & \textcolor{blue}{0.75} & 0.61 & 0.67 & \textcolor{blue}{0.69} & \textcolor{red}{0.77} & \textcolor{blue}{0.73} \\
WhatDoYouSee   & 0.60 & 0.70 & 0.65 & \textcolor{red}{0.75} & 0.53 & 0.62 \\
OTTER          & \textcolor{red}{0.81} & \textcolor{blue}{0.81} & \textcolor{red}{0.81} & \textcolor{red}{0.75} & \textcolor{blue}{0.75} & \textcolor{red}{0.75} \\ \hline
\end{tabular}
\end{table}

CLIP and SigLIP2 adopt dual-encoder architectures that independently encode images and text into a shared embedding space, optimized via large-scale contrastive pre-training. For label prediction, we follow the standard zero-shot classification protocol: the cosine similarity between the input representation and each candidate label embedding is computed, and predictions are obtained by applying a fixed similarity threshold. Specifically, CLIP (ViT-L/14) employs a similarity threshold of 17, while SigLIP2 uses a threshold of 5e-4. These thresholds were selected based on empirical tuning to balance precision and recall.

Qwen2VL represents the class of instruction-tuned MLLMs that integrate a pretrained language backbone with a visual encoder via cross-modal attention. This architecture enables direct conditioning on both modalities for open-ended generation and classification tasks.

WhatDoYouSee is an MLLM-based approach optimized for single-label recognition, which we adapt for multi-label recognition. In our implementation, we select the top-2 candidate labels ranked by model logits and apply a logits threshold of 0.01 to filter low-confidence predictions.

\subsubsection{Evaluation Metrics} Following~\cite{ridnik2023ml,abdelhamed2024you,ali2023clip}, we evaluate OTTER and all baseline models using Precision, Recall, and F1-Score, where higher values indicate better performance. These metrics provide complementary perspectives on model effectiveness in both predefined and open-set label recognition scenarios.

\subsubsection{Implementation Details} All experiments are conducted on a distributed setup of eight NVIDIA GeForce RTX 3090 GPUs (24GB). As illustrated in Fig.~\ref{fig:otter-data}, our training data pipeline incorporates ground-truth open-set labels with a probability of 0.5, enabling the model to learn robust decision boundaries under partial label exposure. We adopt a modular optimization scheme in which the parameters of the image encoder and text encoder are frozen, while all remaining components are fine-tuned. This approach leverages the strong generalization capabilities of large-scale pretrained encoders while reducing computational overhead. The model is trained using the Binary Cross-Entropy loss, a standard choice for multi-label classification tasks due to its ability to handle independent label probabilities. The initial learning rate is set to 1e-3 and scheduled via cosine annealing with warm restarts (T0=100), following the SGDR strategy~\cite{loshchilov2016sgdr}. This scheduling method gradually decays the learning rate following a cosine curve and periodically resets it. We employ a batch size of 64 and train for 200 epochs. All experiments are implemented in PyTorch, with mixed-precision training enabled to optimize GPU memory usage and throughput.

\subsection{Experimental Results}
\subsubsection{Comparison with Baseline}
Across both the Otter and Favorite test sets, OTTER consistently outperforms competing algorithms in overall label prediction accuracy, as reflected by its balanced Precision, Recall, and F1-Score profiles, as shown in Table~\ref{tab:otter_favorite}. Specifically, OTTER achieves an F1 of 0.81 on Otter and 0.75 on Favorite, surpassing the next-best baseline by margins of 0.1 and 0.02, respectively. This advantage extends to fine-grained evaluation by label type: as shown in Table~\ref{tab:otter_favorite_tags}, for open-set labels, OTTER attains near-perfect performance (F1 of 0.99 on Otter, 0.97 on Favorite), markedly higher than all baselines, while maintaining competitive results on predefined labels. Such robustness across both closed and open-set categories suggests that OTTER effectively mitigates the label distribution mismatch problem that often hampers zero-shot and multi-modal classifiers. These findings demonstrate OTTER's capacity to deliver state-of-the-art performance in both conventional and open-vocabulary evaluation scenarios.

\subsubsection{Top Tag Performance}
We measure OTTER's performance on each detailed tags and find that OTTER achieves consistently high accuracy on predefined labels and six most frequent open-set labels, demonstrating robust performance across heterogeneous semantic categories. As shown in Table~\ref{tab:otter_favorite_by_tag}, for open-tags, OTTER attains near-perfect F1-scores in domains such as Epidemic Prevention \& Control (0.99), Public Health (0.98), Laws \& Regulations (0.98), and International Relations (1.00) in Otter dataset. OTTER also reach perfect scores 1.00 for multiple high-frequency categories in Favorite dataset (e.g., Medical Examination Report, Cards \& Certificates, Home-Style Dishes, Winter Delicacies), and sustaining competitive performance in more visually diverse classes such as Culinary Exploration (0.92) and Artificial Intelligence (0.87). This balanced strength across both predefined and open-vocabulary settings suggests that OTTER effectively mitigates the distributional shift and semantic variability challenges that often degrade zero-shot and open-set classification performance.

\begin{table*}[t]
\caption{Performance Comparison on Otter and Favorite Datasets by Tag}
\centering
\vspace{-3mm}
\label{tab:otter_favorite_by_tag}
\setlength{\tabcolsep}{3pt}
\begin{tabular}{
    l|
    >{\centering\arraybackslash}p{0.9cm}
    >{\centering\arraybackslash}p{0.9cm}
    >{\centering\arraybackslash}p{0.9cm}|
    l|
    >{\centering\arraybackslash}p{0.9cm}
    >{\centering\arraybackslash}p{0.9cm}
    >{\centering\arraybackslash}p{0.9cm}
}
\hline
\multicolumn{1}{c|}{Top Tags in Otter} & Prec. & Rec. & F1 & \multicolumn{1}{c|}{Top Tags in Favorite} & Prec. & Rec. & F1 \\ \hline
Learning \& Growth & 0.65 & 0.50 & 0.56 & Learning \& Growth & 0.66 & 0.85 & 0.74 \\
Creative Design & 0.57 & 0.51 & 0.54 & Creative Design & 0.91 & 0.26 & 0.41 \\
Life Moments & 0.76 & 0.71 & 0.73 & Life Moments & 0.67 & 0.79 & 0.72 \\
Tech Frontiers & 0.56 & 0.53 & 0.54 & Tech Frontiers & 0.94 & 0.61 & 0.74 \\
Career \& Business & 0.78 & 0.82 & 0.80 & Career \& Business & 0.63 & 0.91 & 0.74 \\
Sports \& Health & 0.85 & 0.86 & 0.86 & Sports \& Health & 0.82 & 0.80 & 0.81 \\
Epidemic Prevention \& Control & 0.98 & 1.00 & 0.99 & Medical Examination Report & 1.00 & 1.00 & 1.00 \\
Public Health & 0.97 & 1.00 & 0.98 & Artificial Intelligence & 0.77 & 1.00 & 0.87 \\
Laws \& Regulations & 0.97 & 1.00 & 0.98 & Cards \& Certificates & 1.00 & 1.00 & 1.00 \\
Market Analysis & 0.94 & 1.00 & 0.97 & Home-Style Dishes & 1.00 & 1.00 & 1.00 \\
International Relations & 1.00 & 1.00 & 1.00 & Culinary Exploration & 0.86 & 1.00 & 0.92 \\
Legal Cases & 1.00 & 1.00 & 1.00 & Winter Delicacies & 1.00 & 1.00 & 1.00 \\ \hline
\end{tabular}
\end{table*}

\begin{table}[htbp]
\caption{Ablation Study on Training Strategy}
\centering
\vspace{-3mm}
\label{tab:ablation_training_strategy}
\setlength{\tabcolsep}{3pt}
\begin{tabular}{l|
    >{\raggedright\arraybackslash}p{1.8cm}|
    >{\centering\arraybackslash}p{0.9cm}
    >{\centering\arraybackslash}p{0.9cm}
    >{\centering\arraybackslash}p{0.9cm}}
\hline
\multicolumn{2}{c|}{Method} & Prec. & Rec. & F1 \\ \hline
Original                & \checkmark & 0.73 & 0.74 & 0.73 \\
Explain Label Details   & \checkmark\ \checkmark & 0.75 & 0.73 & 0.74 \\
Add OCR Keyword         & \checkmark\ \checkmark\ \checkmark & \textcolor{blue}{0.78} & 0.80 & \textcolor{blue}{0.79} \\
Add Text Title Keyword  & \checkmark\ \checkmark\ \checkmark\ \checkmark & \textcolor{red}{0.81} & \textcolor{blue}{0.81} & \textcolor{red}{0.81} \\
Asymmetric Loss         & \checkmark\ \checkmark\ \checkmark\ \checkmark\ \checkmark & 0.76 & \textcolor{red}{0.83} & \textcolor{blue}{0.79} \\ \hline
\end{tabular}
\end{table}

\subsection{Ablation Study}
\subsubsection{Training Strategy}
To investigate the impact of training strategies on tagging accuracy, we conducted a series of controlled ablation experiments. Starting from the baseline configuration, we progressively enriched the training inputs with semantically informative cues. Specifically, we augmented the label descriptions with additional contextual information: for image-based samples, we incorporated OCR-extracted keywords; for text-based samples, we appended the corresponding document titles. As shown in Table~\ref{tab:ablation_training_strategy}, this incremental enrichment consistently improved the final tagging accuracy, suggesting that richer semantics can enhance the model's discriminative capacity.

From the data distribution perspective, we addressed the inherent label imbalance, which can bias model optimization toward majority classes. We experimented with an Asymmetric Loss formulation, motivated by its reported effectiveness in mitigating imbalance in visual recognition and tagging scenarios. However our ablation result shows this modification did not yield a consistent improvement in tagging accuracy across evaluation splits. Consequently, we retained the Binary Cross-Entropy loss in the final configuration, as it provided more stable convergence without sacrificing performance on minority labels.

\subsubsection{Merging Operator}
As our framework integrates multi-source, multi-modal inputs, it is necessary to fuse the encoded embeddings before passing them to the multi-head attention layer, as in Fig.~\ref{fig:otter-model}. We evaluated four representative merging operators—concatenation (Cat), element-wise maximum (Max), element-wise median (Median), and element-wise addition (Add). As shown in Table~\ref{tab:ablation_merging}, the experimental results indicate that the Add operator achieved the highest overall performance, with a precision of 0.81, recall of 0.81, and F1 score of 0.81. This suggests that additive fusion effectively preserves complementary information from different modalities while maintaining balanced precision-recall trade-offs. In contrast, concatenation yielded competitive recall but slightly lower precision, likely due to the increased dimensionality introducing redundancy and overfitting risk. The Max and Median operators achieved comparable F1 scores, with Median favoring recall but at the cost of reduced precision.

\begin{table}[tp]
\caption{Ablation Study on Merging Operation}
\centering
\vspace{-3mm}
\label{tab:ablation_merging}
\setlength{\tabcolsep}{3pt} % 缩小列间距
\begin{tabular}{l|
    >{\centering\arraybackslash}p{0.9cm}
    >{\centering\arraybackslash}p{0.9cm}
    >{\centering\arraybackslash}p{0.9cm}}
\hline
Operation & Prec. & Rec. & F1 \\ \hline
Cat    & \textcolor{blue}{0.78} & 0.80 & \textcolor{blue}{0.79} \\
Max    & 0.77 & 0.80 & 0.78 \\
Median & 0.76 & \textcolor{red}{0.82} & \textcolor{blue}{0.79} \\
Add    & \textcolor{red}{0.81} & \textcolor{blue}{0.81} & \textcolor{red}{0.81} \\ \hline
\end{tabular}
\end{table}

\section*{Conclusion}
We presented OTTER, a unified open-set multi-label tagging framework that integrates a stable taxonomy of predefined categories with an extensible open-tag layer. By constructing a large-scale, hierarchically structured multi-modal dataset and employing a hybrid annotation pipeline, OTTER achieves both semantic reliability and broad coverage. Our multi-head attention architecture enables precise alignment between multi-modal inputs and heterogeneous label embeddings. Experimental results on two benchmark datasets confirm OTTER's superior performance in both conventional and open-vocabulary scenarios, achieving near-perfect accuracy on open-set labels without sacrificing predefined label quality. We believe OTTER offers a practical and scalable solution for real-world tagging systems, where consistent categorization and personalized expressiveness must coexist, and we anticipate its adoption as a strong baseline for future research in open-set multi-modal classification.

\bibliographystyle{IEEEtran}
% \bibliography{ref}

\appendix
\renewcommand{\thesection}{\Alph{section}}
\renewcommand{\thesubsection}{\thesection-\Alph{subsection}}
\section{Appendix}\label{appendix}
\subsection{Annotation Commands}\label{appendix-A}
For images, textual elements embedded within the visual content—such as OCR characters and descriptive captions—often provide valuable cues for label inference by large models. Accordingly, we pre-extract both the embedded text and the associated captions and include them in the prompt for label assignment. For textual data, filenames frequently carry semantically relevant information that aids the model's classification, hence filename content is likewise incorporated into the labeling input. The prompt structures used for image and text labeling are presented in Listing~\ref{lst:image-prompt} and Listing~\ref{lst:text-prompt}, respectively.

\lstset{
  basicstyle=\ttfamily\small,
  breaklines=true,
  breakatwhitespace=false, % 允许在非空格处断行
  frame=none,
  captionpos=b,
  columns=flexible,
}

\clearpage
\onecolumn

\begin{lstlisting}[caption={Prompt given to ChatGPT for image tag generation. It populates the system field in the ChatGPT API.}, label={lst:image-prompt}]
You are an expert in image tagging. Your task is to assign semantic tags to an image based on its visual content, embedded text, and descriptive caption. The tagging should be open-ended and context-aware. Tags must be accurate and concise: you may assign 0, 1, 2, or 3 tags. Each tag should be no longer than 10 words, and each tag should represent a distinct concept. You may select tags from the predefined list below, or freely generate new tags that match the image content. If suitable tags exist in the predefined list, prefer selecting from them.

Predefined Tags:
Tech Frontiers: Includes all content related to cutting-edge technological inventions, innovations, scientific research progress, future trend forecasting, and applied technology cases—especially those with potential impact on industry or society. Example scenarios include new energy vehicles, artificial intelligence, and digital products.
Career & Business: Covers market analysis, corporate strategy, financial management, human resources, product development, marketing strategy, supply chain management, customer relations, entrepreneurship, and international business. Example scenarios include annual reports, financial statements, and industry trends.
Creative Design: Focuses on creative thinking, design theory, art appreciation, design case studies, and trend analysis. Intended to stimulate creativity and provide design inspiration. Example scenarios include fashion styling and aesthetic design.
Life Moments: Captures moments from personal life, including family activities, travel experiences, culinary exploration, holiday celebrations, and personal achievements. Reflects the richness and diversity of daily living. Example scenarios include leisure activities, holidays, schedules, computer games, passwords, ID cards, and WeChat screenshots.
Learning & Growth: Encompasses educational information, learning resources, skills training, personal development strategies, career planning, and self-improvement. Aims to support lifelong learning and personal advancement. Example scenarios include workplace development and error notebooks.
Sports & Health: Includes healthy eating, fitness, mental health, disease prevention, healthy habits, and medical information. Promotes healthy lifestyles and improved quality of life. Example scenarios include medical checkup reports, fitness activities, and disease treatment.

Example 1 
Image Input: An image showing traditional Chinese massage therapy for treating pharyngitis 
OCR Text: Sumu’s original intention 1/5 X relief throat inflammation pain golden technique tips Shaoshang acupoint midpoint Yuji acupoint Shangyang acupoint medial ankle tip Zhaohai acupoint… These acupoints are considered highly effective in treating throat inflammation.
Caption: This image illustrates a therapy called “Golden Technique Tips,” featuring five hand and foot acupoints aimed at relieving cough and throat pain. It includes concise textual explanations and diagrams emphasizing the effectiveness of these acupoints. 
Your Output: Sports & Health, Home Remedies, Massage

Example 2 
Image Input: A scene from the launch event of the Wenxin large-scale model 
OCR Text: Extensive list of model names, APIs, deployment tools, and industry applications
Caption: At a conference, a speaker stands on stage in front of a slide showcasing Wenxin’s industrial-grade knowledge-enhanced large model. The content covers multiple model categories and applications, including NLP and CV. The audience listens attentively. 
Your Output: Tech Frontiers, Artificial Intelligence, Large Models

Please output only the selected or generated tags. 

Now, I will give you an image. 

Image Input: Please refer to the image content. 
OCR Text: {ocr_text} 
Caption: {caption_text} 
Your Output:
\end{lstlisting}

\clearpage
\begin{lstlisting}[caption={Prompt given to ChatGPT for text tag generation. It populates the system field in the ChatGPT API.}, label={lst:text-prompt}]
You are an expert in text tagging. Your task is to assign semantic tags to an image based on the document’s filename and the content of its text paragraphs. The tagging should be open-ended and context-aware. Tags must be accurate and concise: you may assign 0, 1, 2, or 3 tags. Each tag should be no longer than 10 words, and each tag should represent a distinct concept. You may select tags from the predefined list below, or freely generate new tags that match the text content. If suitable tags exist in the predefined list, prefer selecting from them.

Predefined Tags:
Tech Frontiers: Includes all content related to cutting-edge technological inventions, innovations, scientific research progress, future trend forecasting, and applied technology cases—especially those with potential impact on industry or society. Example scenarios include new energy vehicles, artificial intelligence, and digital products.
Career & Business: Covers market analysis, corporate strategy, financial management, human resources, product development, marketing strategy, supply chain management, customer relations, entrepreneurship, and international business. Example scenarios include annual reports, financial statements, and industry trends.
Creative Design: Focuses on creative thinking, design theory, art appreciation, design case studies, and trend analysis. Intended to stimulate creativity and provide design inspiration. Example scenarios include fashion styling and aesthetic design.
Life Moments: Captures moments from personal life, including family activities, travel experiences, culinary exploration, holiday celebrations, and personal achievements. Reflects the richness and diversity of daily living. Example scenarios include leisure activities, holidays, schedules, computer games, passwords, ID cards, and WeChat screenshots.
Learning & Growth: Encompasses educational information, learning resources, skills training, personal development strategies, career planning, and self-improvement. Aims to support lifelong learning and personal advancement. Example scenarios include workplace development and error notebooks.
Sports & Health: Includes healthy eating, fitness, mental health, disease prevention, healthy habits, and medical information. Promotes healthy lifestyles and improved quality of life. Example scenarios include medical checkup reports, fitness activities, and disease treatment.

Example 1 
Filename: Airport Transportation FAQ.docx 
Text Input: Detailed description of electric cart services, application procedures, and ground transportation options provided by China Southern Airlines at Guangzhou Baiyun Airport.
Your Output: Life Moments, Airport Transportation, Shuttle Services

Example 2 
Filename: Medical Checkup Report.pdf 
Text Input: Comprehensive medical examination results including blood pressure, vision, blood tests, ultrasound findings, and lifestyle recommendations.
Your Output: Sports & Health, Medical Report

Please output only the selected or generated tags. 

Now, I will give you a filename: {title}
Text Input: {text}
Your Output:
\end{lstlisting}

\end{document}